\documentclass[singlecolumn]{starpaper}

\usepackage{hyperref}
\usepackage{url}
\usepackage{graphicx}
\usepackage{wrapfig}
\usepackage{booktabs}
\usepackage{tabularx}
\usepackage{array}
\usepackage{multirow}
\usepackage{tcolorbox}
\usepackage{amsmath}
\usepackage{xspace}
\usepackage{flafter} 
\usepackage{placeins}

\title{\huge God Does Not Play Dice: Enforcing Batch Invariance in Agentic Kernel Generation}

\author{Youwei Bao\textsuperscript{1,\textdagger}\hspace{1.5em}%
Chengpeng Wang\textsuperscript{2}\hspace{1.5em}%
Simin Chen\textsuperscript{1,*}}

\staraffiliation{\textsuperscript{1}STAR Lab, George Mason University\qquad
\textsuperscript{2}National University of Singapore}
\staremails{}
\starshorttitle{Batch Invariance in Agentic Kernel Generation}
\startrack{ARTIFICIAL INTELLIGENCE / MACHINE LEARNING}
\starstatus{PREPRINT}
\starversion{v1}
\starkeywords{LLM Agents, GPU Kernel Optimization, Code Generation, Batch Invariance, Reproducibility}
\setcitestyle{numbers,square,semicolon}
\stargmulogo{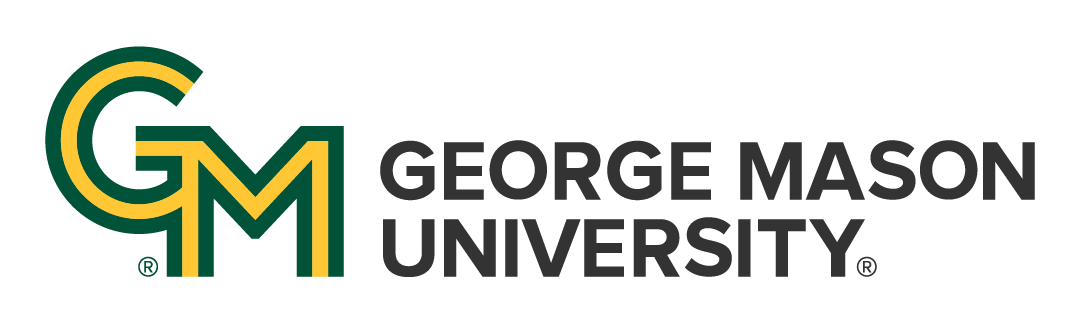}
\starpartner{National University of Singapore}{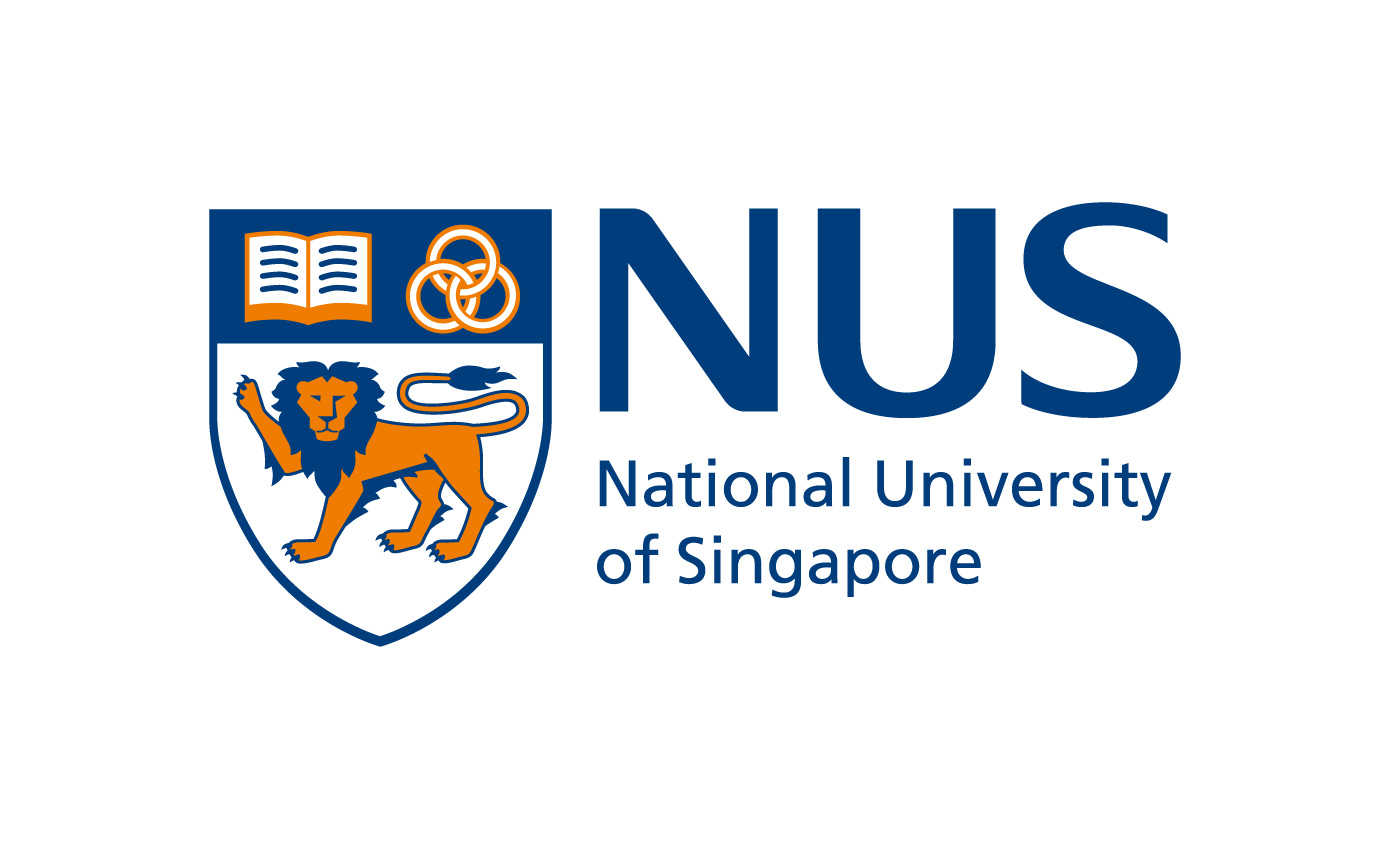}
\makeatletter
\renewcommand{\star@brandrow}{%
  \noindent\begin{minipage}[c][64pt][c]{.40\textwidth}
    \includegraphics[width=\linewidth]{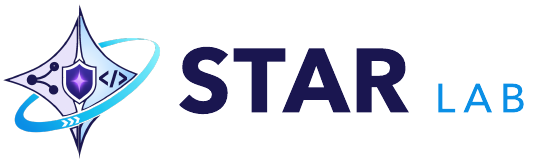}%
  \end{minipage}%
  \hspace{.02\textwidth}%
  {\color{STARLine}\raisebox{-22pt}{\rule{.4pt}{48pt}}}%
  \hfill%
  \begin{minipage}[c][64pt][c]{.209\textwidth}
    \includegraphics[width=\linewidth]{assets/gmu-logo.png}%
  \end{minipage}\hspace{.055\textwidth}%
  \begin{minipage}[c][64pt][c]{.18\textwidth}\raggedleft
    \sbox0{\includegraphics{assets/nus-logo-color.jpg}}%
    \includegraphics[trim=.14\wd0 .22\ht0 .14\wd0 .21\ht0,clip,height=26.4pt]{assets/nus-logo-color.jpg}%
  \end{minipage}\par}
\renewcommand{\starresources}{%
  \par\vspace{6pt}\noindent\begin{minipage}{\linewidth}\centering
  \sffamily\fontsize{10}{13.5}\selectfont\color{STARPrimary}%
  \star@resourceicon{1}\hspace{4pt}\textbf{GitHub:}\hspace{4pt}\href{https://github.com/STAR-Computing/BIEvolve}{https://github.com/STAR-Computing/BIEvolve}\par
  \end{minipage}\par}
\renewenvironment{abstract}{%
  \par\begingroup\setlength{\parindent}{0pt}%
  {\sffamily\bfseries\small\color{STARForest}ABSTRACT\par}\vspace{4pt}%
  \small\ignorespaces
}{%
  \par\starresources\vspace{6pt}%
  {\sffamily\fontsize{8}{10}\selectfont
    \textbf{Keywords}\quad\star@keywords\par}%
  \endgroup\vspace{8pt}}
\makeatother
\titleformat{\paragraph}[runin]{\bfseries}{\theparagraph}{0.5em}{}
\def \tool{\textit{BIEvolve}\xspace}

\begin{document}

\maketitle
\begingroup
\renewcommand{\thefootnote}{\textdagger}%
\footnotetext[1]{This work is done during the internship in \href{https://star-computing.github.io/}{STAR Lab}, George Mason University.\quad
\textsuperscript{*} Corresponding author.}
\endgroup

\begin{abstract}
Large language model (LLM) serving systems dynamically batch user requests for GPU execution, so the same request may be processed alongside different requests across runs. If the GPU kernels lack batch invariance (BI), this changing batch context can alter the numerical output for an unchanged input, undermining reproducibility.
Existing kernel generation agents focus on correctness and efficiency but \emph{overlook BI}.
We find 19 confirmed BI violations among 83 batch-applicable released artifacts.
To address this limitation, we propose \tool, which combines contract-guided static dependency analysis with dynamic validation of bitwise consistency across batch contexts.
\tool is designed as a plug-in component for general kernel agents, allowing their original optimization capabilities to be reused while adding BI enforcement. We integrate it as successive acceptance gates in the optimization loop, rejecting structurally risky or dynamically non-invariant candidates before performance benchmarking and returning feedback to the agent. 
Experiments across three agent frameworks under baseline and BI-aware generation on real world LLM serving tasks show that \tool improves the BI pass rate from 67.6\% to 91.9\% while retaining 91.8\% of the baseline geometric mean speedup. These results demonstrate that \tool can be incorporated into kernel optimization by agents as an additional reliability objective alongside correctness and speed.

\end{abstract}

\section{Introduction}

Large language model (LLM) agents are increasingly used to generate and optimize high performance GPU kernels~\citep{dong2026stark,cao2026ksearch, yu2026towards, cao2026k}. Existing kernel agents are typically evaluated along two axes: functional correctness and execution efficiency. A generated kernel is considered successful if it matches a reference implementation on test inputs while achieving lower latency, as exemplified by  KernelBench~\citep{ouyang2025kernelbench}. However, this evaluation paradigm overlooks a critical reliability requirement: \textit{whether an unchanged input produces consistent numerical results as its execution context changes.}

\begin{figure}[htbp]
\centering
\includegraphics[width=\linewidth]{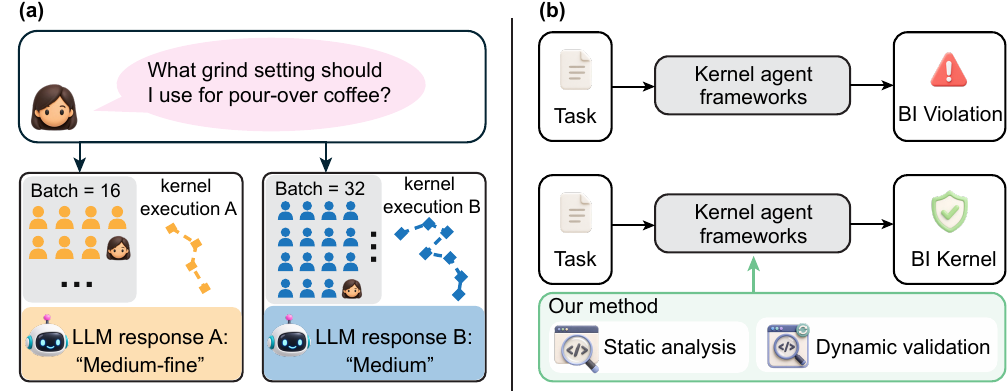} 
\caption{Batch dependence and the \tool workflow. (a) An illustrative request can receive different outputs when its batch context changes. (b) A DSL-guided static analysis module rejects forbidden dependencies specified by contracts, while dynamic validation checks realized behavior across batch contexts.} 
\label{fig:bi_overview}
\end{figure}

We focus on such overlooked reliability property, \emph{batch invariance} (BI). For workloads in which samples are semantically independent, BI requires that an unchanged sample produce exactly the same output regardless of batch size or batch composition~\citep{he2025nondeterminism,vllm2026batchinvariance}. A BI violation occurs when changing the batch context changes that sample's output, as illustrated in {Figure~\ref{fig:bi_overview}(a)}. Existing generated kernels are not necessarily BI because changing the batch context can alter low-level execution decisions, such as kernel tiling, parallel decomposition, and reduction order. Since floating point arithmetic is non-associative, these implementation changes can produce different results even with identical model weights and deterministic sampling settings~\citep{he2025nondeterminism,gond2026llm42, chen2025compiler}. Thus, a kernel may pass conventional correctness tests while still exhibiting batch-dependent behavior, undermining reproducibility, reliability and security~\citep{ouyang2025empirical, chen2025compiler}.

Evidence from released agent-generated kernels confirms that the problem occurs in practice (Table~\ref{tab:released_audit}). Among 131 released artifacts, 83 have a meaningful semantic batch dimension, and 19 of these 83 (22.9\%) contain confirmed BI violations.

\begin{wraptable}{r}{0.54\textwidth}
\vspace{-0.5\baselineskip}
\caption{BI evidence from released kernel artifacts.}
\label{tab:released_audit}
\centering
\footnotesize
\setlength{\tabcolsep}{2.5pt}
\renewcommand{\arraystretch}{1.03}
\begin{tabularx}{\linewidth}{@{}>{\raggedright\arraybackslash}Xccc@{}}
\toprule
\shortstack[l]{Artifact source /\\ benchmark} & \shortstack{Batch-\\applicable /\\ released} & \shortstack{Confirmed\\ BI violations} & \shortstack{BI pass\\ (\%)} \\
\midrule
KernelBench.com / KernelBench.com & 79 / 126 & 18 & 77.2 \\
K-Search / FlashInfer-Bench & 4 / 5 & 1 & 75.0 \\
\midrule
\textbf{Total} & \textbf{83 / 131} & \textbf{19} & \textbf{77.1} \\
\bottomrule
\end{tabularx}
\vspace{-0.35\baselineskip}
\end{wraptable}

BI is particularly important in modern LLM systems because batch contexts change continuously in practice. Dynamic and continuous batching repeatedly regroup user prompt request, so the same request may be executed under different batch configurations across runs~\citep{kwon2023pagedattention}. Even small numerical deviations can propagate through autoregressive decoding and alter the final outputs. Enforcing BI therefore improves reproducibility and predictable serving, facilitates debugging by making divergent operations easier to localize~\citep{zhou2026opguard}, and can reduce batch-dependent rollout, training mismatch in LLM reinforcement learning~\citep{yao2025mismatch,zhang2026deterministic,zhong2026mismatch}.

Existing work has introduced BI kernels and deterministic execution modes for specific inference systems~\citep{he2025nondeterminism,vllm2026batchinvariance}. However, these methods rely on specialized, system-specific changes and cover only a limited set of kernels, making them hard to scale to agent-generated kernels. For example, \citet{thinkingmachines2025ops} reports replacing only \textbf{four} PyTorch operators, providing limited coverage of the operator space. Moreover, recovering performance after enforcing BI execution still requires substantial kernel engineering, motivating an optimization agent that can search for efficient invariant implementations across operators and hardware. At the same time, replacing existing kernel agents would lose the strong optimization capabilities they already provide. This motivates a more general approach: 

\begin{center}
\begin{tcolorbox}[colback=gray!6,
colframe=gray,
width=0.99\textwidth,
arc=1mm,
boxrule=0.9pt]
\textit{Adding batch invariance as a lightweight plug-in to existing kernel generation agents, while reusing their original optimization pipelines.}
\end{tcolorbox}
\end{center}

{Figure~\ref{fig:bi_overview} shows the working mechnisms of \tool, which combines} a domain specific language (DSL)-guided static analysis module with dynamic validation module. The DSL specifies BI-relevant task semantics, and the static analysis module consumes this specification to detect forbidden dependencies from declared batch symbols to protected parameters. The dynamic validation module compares an unchanged target sample across batch contexts. Candidates pass through DSL-guided static analysis, reference correctness testing, and dynamic validation before performance benchmarking. Rejected candidates receive actionable diagnostics that guide subsequent optimization attempts.

We plug \tool into three representative LLM-based kernel optimization agents and evaluate it, evolving existing kernel agents to generate BI kernels. Across controlled agent reruns, \tool improves the BI pass rate from 67.6\% to 91.9\% while retaining 91.8\% of the baseline geometric mean speedup on tasks covering diverse kernel families and subtypes. We report tolerance-based reference correctness separately from exact BI and analyze the complementary violations detected by the static analysis module and dynamic validation module.

Our contributions are:
\begin{itemize}
    \item \textbf{Problem Novelty.} We identify BI as an overlooked reliability requirement in LLM-driven GPU kernel generation and optimization.
    \item {\textbf{Technique Novelty.} We introduce \tool, a plug-in for general kernel generation and optimization agents, combining a DSL-guided static analysis module with dynamic validation across semantically matched batch contexts.}
    \item \textbf{ Empirical Evaluation.} We integrate \tool into three agent pipelines and evaluate \tool on a task suite drawn from KernelBench.com and FlashInfer Bench workloads.
\end{itemize}

To our knowledge, this is the first work to make BI an acceptance criterion during LLM-driven GPU kernel optimization.

\section{Background and Problem Statement}


\subsection{GPU Kernel Execution and Nondeterminism}

A GPU executes a kernel by launching a grid of thread blocks, each containing multiple threads. To improve hardware utilization, GPU operator implementations often specialize execution to the input shape through library dispatch or shape-dependent kernel logic. Numerical outputs can therefore vary across executions for several distinct reasons, depending on what changes between them. Run-to-run nondeterminism can arise even for the same input and execution context, originating from unordered floating point atomics, although such atomics are uncommon in modern LLM forward inference. Numerical differences can also arise when the hardware or software environment changes~\cite{chen2023dycl, chen2025compiler, wang2026hidden, moller2026hardware, rahaman2026ispy, loose2026floatdoor}. These sources are distinct from batch-dependent variation, where the target input and execution environment remain unchanged but its surrounding batch context changes.

We focus on batch-dependent variation because batch contexts naturally change in modern LLM serving~\cite{kwon2023efficient,pochelu2021efficient, pang2023efficient}. Dynamic batching and scheduling can alter the batch size, composition, and associated workload shapes seen by a kernel, even when the target request itself is unchanged. Consequently, changing the batch context can select a different kernel variant, alter tile dimensions, or change how work is partitioned across thread blocks, including the use of split reductions.

For reductions such as summation, matrix multiplication, and normalization, these threads compute partial results in parallel and subsequently combine them. Since floating-point addition is not associative, rounding can cause $(a+b)+c$ to differ from $a+(b+c)$. The numerical result can therefore depend on how a reduction is partitioned and in which order its partial results are accumulated~\citep{rathnasuriya2025investigation, chen2025compiler}. Changing the batch context can consequently change the reduction order, and thus the output bits for a target sample, even though the sample and its mathematical computation remain unchanged \citep{nvidia2025floatingpoint,he2025nondeterminism, chen2025compiler}.

\begin{figure}[!ht]
    \centering
    \includegraphics[width=0.96\linewidth]{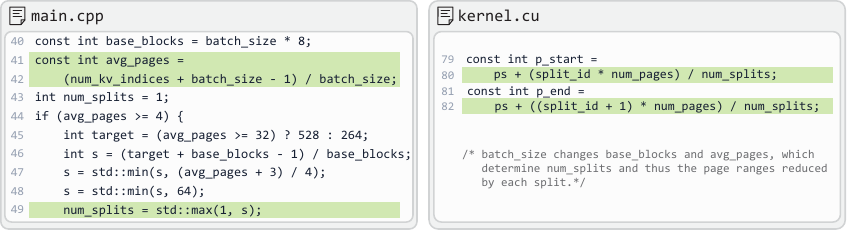}
    \caption{Representative kernel code patterns that introduce batch variance.}
    \label{fig:batch_variance_code}
\end{figure}

In attention, for example, changes in workload shape can alter tiling or Split-KV strategies and thereby change the floating point reduction order. Prior studies therefore identify batch-dependent changes in floating point reduction and decomposition as a central source of batch variance in LLM kernels \citep{he2025nondeterminism,cankaya2026bitexact}. In Figure~\ref{fig:batch_variance_code}, we provide a concrete code example identified in our study of a batch dependent split/reduction: \texttt{batch\_size} changes \texttt{base\_blocks} and \texttt{avg\_pages}, which determine \texttt{num\_splits} and thus the page ranges reduced by each split.

\FloatBarrier

\subsection{Batch Invariance}

Unless otherwise stated, all executions considered below use the same hardware and software enviroment. Let $K(B)$ denote the output of a candidate kernel $K$ on a batch $B$, and let $\mathcal{X}$ denote the input domain. For each target sample $x \in \mathcal{X}$, let $\mathcal{C}(x)$ be the set of admissible batch contexts that preserve $x$ and its reference semantics while allowing the batch size, composition, and position of $x$ to vary. A context $c \in \mathcal{C}(x)$ produces a batch $B_c$ in which $x$ appears at position $i_c$. We say that $K$ is batch invariant over $\mathcal{X}$ if
$\forall x \in \mathcal{X}$ and $\forall c,c' \in \mathcal{C}(x)$,
$K(B_c)_{i_c} \equiv_{\mathrm{bit}} K(B_{c'})_{i_{c'}}$.

Thus, the output associated with a preserved target sample must be bitwise identical across all admissible batch contexts. This definition treats batch invariance as an umbrella property encompassing invariance to batch size, batch composition, and batch position; restricting $\mathcal{C}(x)$ to changes along one dimension yields the corresponding narrower property.

\paragraph{Problem definition.} Given a kernel task, its reference implementation, and a set of admissible batch contexts, our goal is to generate an optimized kernel that (i) satisfies the task's numerical correctness criterion, (ii) is BI across admissible contexts under a fixed hardware and software environment, and (iii) achieves high kernel execution performance.

\section{Methodology}
\label{sec:methodology}
\begin{figure}[!ht]
\centering
\includegraphics[width=\linewidth]{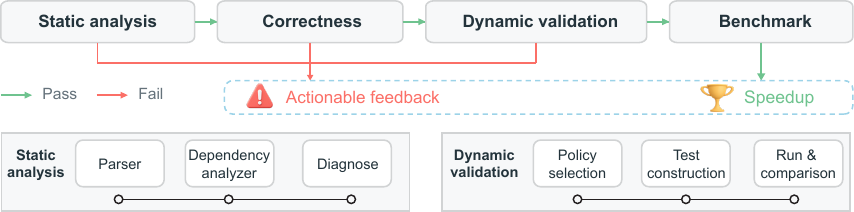}
\caption{\tool candidate evaluation workflow. BI failures produce actionable feedback, while accepted candidates proceed to speedup measurement.}
\label{fig:bievolve_pipeline}
\end{figure}

\tool is designed as a plug-in workflow for existing kernel agents. As illustrated in Figure~\ref{fig:bi_overview}, it combines a DSL-guided static analysis module with dynamic validation module, while retaining each framework's native correctness test and performance benchmark. Figure~\ref{fig:bievolve_pipeline} provides a detailed view of \tool. Both modules are driven by domain knowledge, the design rules could also be automatically extracted by an agent from representative examples. The static module uses contracts that specify forbidden batch dependencies into protected parameters; the dynamic module tests whether a preserved sample produces bitwise identical outputs across valid batch contexts. When a candidate fails either module, \tool returns actionable diagnostics through the agent's existing feedback channel to guide the next optimization round.

\FloatBarrier

\subsection{DSL-Guided Static Analysis Module}
\label{sec:dynamic_test}

\paragraph{BI contract DSL.} The static analysis module detects source level dependencies that are known to threaten BI before executing a generated kernel. It takes two inputs: task metadata describing the task semantics and the generated source bundle. The metadata is used to select the applicable BI contract, while the source code is parsed into a representation suitable for dependency analysis. The analyzer then checks whether any protected kernel parameter depends on the batch context. The module outputs either a pass decision or an actionable diagnosis identifying the violating dependency.

The contract DSL specifies two elements: batch context variables, including batch size and batch-derived workload quantities, and protected parameters, whose values must remain independent of the batch context variables. Contracts are organized hierarchically as global, kernel family, and subtype rules; task metadata resolves the applicable contracts, allowing a new task to inherit protections from its semantic family. Aliases map implementation specific variable names to canonical protected parameters. The simplified DSL grammar is shown below in Figure~\ref{fig:bi-dsl}.

\begin{figure}[!t]
\centering
\begin{minipage}{0.78\linewidth}
\centering
\begingroup\footnotesize
\setlength{\tabcolsep}{1.5pt}
\renewcommand{\arraystretch}{0.94}
\newcommand{\ntdsl}[1]{\ensuremath{\langle\textit{#1}\rangle}}
\begin{tabular}{@{}r@{\;}c@{\;}l@{}}
\ntdsl{contract} & $::=$ & \textnormal{"kernel"} \ntdsl{id} \ntdsl{inherit} \\
& & \textnormal{"\{"} \ntdsl{axis} \ntdsl{params} \textnormal{"\}"} \\
\ntdsl{inherit} & $::=$ & $\varepsilon$ \\
& & $\mid$ \textnormal{"extends"} \ntdsl{names} \\
\ntdsl{axis} & $::=$ & \textnormal{"invariant\_axis"} \\
& & \textnormal{"batch" "symbols"} \\
& & \textnormal{"["} \ntdsl{names} \textnormal{"]" ";"} \\
\ntdsl{params} & $::=$ & $\varepsilon$ \\
& & $\mid$ \ntdsl{parameter} \ntdsl{params} \\
\ntdsl{parameter} & $::=$ & \textnormal{"parameter"} \ntdsl{id} \\
& & \textnormal{"\{"} \ntdsl{fields} \textnormal{"\}"} \\
\ntdsl{fields} & $::=$ & \ntdsl{field} \\
& & $\mid$ \ntdsl{field} \ntdsl{fields} \\
\ntdsl{field} & $::=$ & \ntdsl{semantic} \\
& & $\mid$ \ntdsl{policy} \\
& & $\mid$ \ntdsl{aliases} \\
\ntdsl{semantic} & $::=$ & \textnormal{"semantic"} \ntdsl{sem} \textnormal{";"} \\
\ntdsl{sem} & $::=$ & \textnormal{"split\_structure"} \\
& & $\mid$ \textnormal{"recurrence\_structure"} \\
& & $\mid$ \textnormal{"rng\_stream"} \\
\ntdsl{policy} & $::=$ & \textnormal{"policy"} \ntdsl{policy\_kind} \textnormal{";"} \\
\ntdsl{policy\_kind} & $::=$ & \textnormal{"fixed\_across\_batch"} \\
& & $\mid$ \textnormal{"may\_depend\_on" "batch"} \\
& & $\mid$ \textnormal{"free"} \\
\ntdsl{aliases} & $::=$ & \textnormal{"aliases" "["} \ntdsl{names} \textnormal{"]" ";"} \\
\ntdsl{names} & $::=$ & \ntdsl{id} \\
& & $\mid$ \ntdsl{id} \textnormal{","} \ntdsl{names} \\
\ntdsl{id} & $::=$ & \textnormal{[A-Za-z\_][A-Za-z0-9\_]*}
\end{tabular}
\par\vspace{1pt}
{\scriptsize\raggedright\emph{Semantic constraints: inheritance is resolved before analysis; aliases denote the same parameter; policies respectively forbid batch dependence, permit dependence on batch, or leave dependence unrestricted.}\par}
\endgroup
\caption{Simplified Backus-Naur Form (BNF) grammar of the BI contract DSL.}
\label{fig:bi-dsl}
\end{minipage}
\end{figure}
The current DSL protects key mechanisms inclduing: split/reduction structure (\texttt{NUM\_SPLITS}), recurrence decomposition (\texttt{RECURRENCE\_SEGMENTS}), and random stream state (\texttt{RNG\_SEED}).

\paragraph{Source parsing.} Given task metadata, the contract parser resolves the applicable global, family, and subtype contracts and exposes their declared batch sources and protected parameters to the analyzer. In parallel, the source parser processes the complete generated source bundle. Python and Triton code are parsed using abstract syntax tree (AST)-based analysis that tracks dependencies through argument passing, container accesses, control flow, and a restricted set of same file helper functions. C/C++/CUDA code is handled with targeted dependency patterns, and embedded generated native code is included in the same analysis. The resulting representation provides the dependency information required by the subsequent analyzer.

\paragraph{Contract-guided dependency analysis.} The analyzer treats contract-declared batch symbols and their derived global quantities as dependency sources, and protected parameters as sinks. It propagates both data and control dependence through the parsed program. A candidate is rejected if a protected parameter is assigned directly or transitively from batch context, or if its value is selected through a batch dependent conditional or loop. Thus, the contract determines what must remain batch independent, while the analyzer determines whether the generated implementation satisfies that requirement. Dependencies that cannot be reliably resolved statically, such as opaque runtime dispatch or autotuning behavior, are left to the dynamic validation module.

\paragraph{Diagnostic feedback.} For a rejected candidate, the analyzer converts the detected dependency into compact feedback for the kernel agent. It reports the protected mechanism, source code location, and dependency path from the batch source to the protected parameter, together with mechanism specific repair guidance.
\subsection{Dynamic Validation Module}
\label{sec:dynamic_test}

\paragraph{Test policy selection.} The dynamic validation module tests whether an executable candidate produces the same output for an unchanged sample under different batch contexts. Its input is a reference-correct candidate together with task metadata, and its output is either a pass decision or a diagnostic describing the first detected cross batch mismatch. Task metadata selects a predefined batch test policy according to the kernel family or subtype. General kernels use less batch contexts, while higher risk families such as attention use additional contexts to exercise more execution boundaries.

\paragraph{Cross batch comparison.} For each selected batch context, the test harness constructs a valid workload around the same target sample. The target sample is first executed on the baseline batch context, and the target output is recorded. It is then executed under each larger batch context, after which the corresponding target output is extracted and compared bitwise with the baseline. BI requires identical structure, shape, dtype, and numerical values. Testing stops at the first mismatch.

\paragraph{Feedback.} If all tested contexts produce identical target outputs, the candidate passes dynamic validation. Otherwise, dynamic module reports the failing batch context through the agent's existing feedback path.

\subsection{Plug into the Agent Loop}
\label{sec:agent_integration}

\paragraph{Candidate evaluation pipeline.} \tool is integrated into the existing candidate evaluation pipeline rather than modifying the agent's generation or search algorithm. A lightweight framework adapter packages the information required by the static and dynamic modules, inserts both modules into the candidate evaluation pipeline, and converts BI failures into the framework's original feedback format. At each optimization round, a generated candidate passes through four sequential stages: (1) static BI analysis, (2) reference correctness test, (3) dynamic BI validation, and (4) performance benchmarking. The static analysis first removes candidates with recognized BI violation patterns. The framework's native correctness test then filters incorrect implementations. Dynamic validation checks the remaining candidates across batch contexts, and performance is measured only after both correctness and \tool are passed. This ordering prevents unnecessary execution of statically invalid candidates and avoids interpreting correctness failures as BI violations.

\paragraph{Framework adapter.} For each agent framework, the adapter supplies the information needed for the BI modules: the generated source and task metadata for static analysis, the test harness and task metadata for dynamic validation. The framework's original correctness criterion, benchmark procedure, prompt, candidate selection policy, and search state remain unchanged. Thus, the \tool can be attached to agents with different optimization strategies.

In addition, the adapter routes the outputs of the BI modules. If a candidate passes, the evaluation pipeline proceeds. If either BI module rejects the candidate, subsequent evaluation stages are skipped, and the diagnostic is recorded through the framework's existing evaluation history or log and inserted into its feedback path. This allows subsequent optimization rounds to condition on previous BI failures and revise the implementation accordingly.

\section{Experiments}
\label{sec:experiments}

In this section, we evaluate whether \tool reduces BI violations in agent-generated kernels while preserving ordinary correctness and useful performance. The evaluation combines controlled kernel generation with three agent frameworks, subgroup performance analysis, and an ablation of the \tool components.
\subsection{Experimental Setup}

\paragraph{Datasets.} Our controlled kernel generation experiment uses a compact, risk-focused suite drawn from the hard tasks on KernelBench.com \citet{arledge2026kernelbench} and the FlashInfer trace and contest workloads described by \citet{xing2026flashinferbench} and \citet{flashinfer2026contest}. We select batch applicable kernels spanning attention, linear attention, mixture-of-experts (MoE), normalization, and sampling. Attention covers GQA, MLA, and DSA across representative prefill and decode settings, while linear attention includes KDA and Gated DeltaNet variants.

We emphasize attention and linear attention because their execution commonly involves batch dependent partitioning, algorithm selection, and reductions, while the remaining families broaden coverage beyond these higher risk regimes.

\paragraph{Agent frameworks.} We evaluate K-Search from \citet{cao2026ksearch} and the Iterative Agent and Evolve Agent from \citet{flashinfer2026agents} to cover three representative search paradigms in feedback-driven kernel generation. The Iterative Agent refines a current implementation along a single trajectory; the Evolve Agent explores multiple implementations through recent and elite candidate pools; and K-Search guides implementation generation using a co-evolving world model that separates planning from code instantiation.

These frameworks differ in how they retain search history, select candidates for improvement, and explore alternative designs. Together, they provide broad architectural coverage for evaluating whether \tool generalizes beyond a particular refinement policy or search state representation.

For each framework, we compare vanilla generation with a rerun using \tool as described in Section~\ref{sec:methodology}. We use 20 optimization turns for K-Search and 25 turns for the Iterative Agent and Evolve Agent, keeping the turn budget unchanged between vanilla and BI-aware generation within each framework. Controlled generation uses Gemini 3.7 Flash for all three frameworks on an NVIDIA H100 NVL GPU.

\paragraph{Evaluation.} We evaluate the selected kernels using the same protocol across generation configurations. Ordinary correctness compares each kernel with its task reference using the prescribed numerical tolerance. We report these outcomes separately, since tolerance-based agreement with a reference does not imply BI.

Performance is measured as speedup relative to the corresponding task reference and summarized using geometric means. Performance retention is the ratio of the BI-aware and vanilla geometric mean speedups over the same tasks.

\subsection{Main Results}
\label{sec:main_results}

\paragraph{Controlled kernel generation.} We report the final-solution BI pass rate, defined as the fraction of correctness-passing kernels with no confirmed BI violation, and the BI intervention rate, defined as the fraction of tasks for which \tool rejects at least one generated candidate during optimization. Correctness is evaluated using the task-specific reference tolerance. Table~\ref{tab:controlled_generation} compares baseline and BI-aware generation under these metrics. GM denotes geometric mean. Speedup retained is reported once per framework for the BI-aware/baseline comparison. Retained speedup is $100\times\mathrm{GM}_{\mathrm{BI\text{-}aware}}/\mathrm{GM}_{\mathrm{baseline}}$, using the same tasks and reference implementations.

In the K-Search study, \tool increases the BI pass rate from 75.7\% to 97.4\% under the final solution evaluation, while ordinary correctness changes from 94.9\% to 97.4\%. For the Iterative Agent and Evolve Agent, BI pass rates change from 48.6\% to 89.2\% and from 78.4\% to 89.2\%, respectively, while ordinary correctness remains 94.9\% in both conditions for both frameworks.

The BI intervention rate is highest for the Iterative Agent at 43.6\%, compared with 23.1\% for K-Search and 25.6\% for Evolve Agent, indicating that the framework with the weakest baseline BI pass rate required intervention on substantially more tasks and also exhibited the largest final BI pass rate improvement. Notably, intervention rates may exceed the corresponding final BI pass gains, indicating that \tool also filters BI violating intermediate candidates on tasks whose baseline final solution is BI.

\begin{table}[!htb]
\caption{Controlled kernel generation results across agent frameworks.}

\label{tab:controlled_generation}
\centering
\setlength{\tabcolsep}{3pt}
\renewcommand{\arraystretch}{1.12}
\begin{tabular*}{\linewidth}{@{\extracolsep{\fill}}llccccc@{}}
\toprule
\raisebox{0.5\baselineskip}{\shortstack[l]{Framework}} &
\shortstack[l]{Generation\\condition} &
\shortstack{BI pass\\(\%)} &
\shortstack{BI intervention\\(\%)} &
\shortstack{GM speedup\\($\times$)} &
\shortstack{Speedup\\retained (\%)} &
\shortstack{Correctness\\pass (\%)} \\
\midrule

\multirow{2}{*}{\shortstack[l]{K-Search}}
& Baseline & 75.7 & - & 40.94 & \multirow{2}{*}{86.4} & 94.9 \\
& BI-aware & 97.4 & 23.1 & 35.39 & & 97.4 \\

\midrule
\multirow{2}{*}{Iterative Agent}
& Baseline & 48.6 & - & 25.84 & \multirow{2}{*}{87.7} & 94.9 \\
& BI-aware & 89.2 & 43.6 & 22.66 & & 94.9 \\

\midrule
\multirow{2}{*}{Evolve Agent}
& Baseline & 78.4 & - & 22.85 & \multirow{2}{*}{102.0} & 94.9 \\
& BI-aware & 89.2 & 25.6 & 23.31 & & 94.9 \\

\midrule
\multirow{2}{*}{Overall}
& Baseline & 67.6 & - & 28.91 & \multirow{2}{*}{91.8} & 94.9 \\
& BI-aware & 91.9 & 30.8 & 26.54 & & 95.7 \\

\bottomrule
\end{tabular*}
\par\smallskip
\end{table}

\FloatBarrier

\setcounter{figure}{4}

\paragraph{Performance retention.} For K-Search, the BI-aware rerun achieves a geometric mean speedup of 35.39$\times$, compared with 40.94$\times$ for vanilla generation, retaining 86.4\% of the vanilla speedup. The corresponding retention for the Iterative Agent and Evolve Agent is 87.7\% and 102.0\%.

\begin{figure}[!htbp]
\centering
\includegraphics[width=0.70\linewidth]{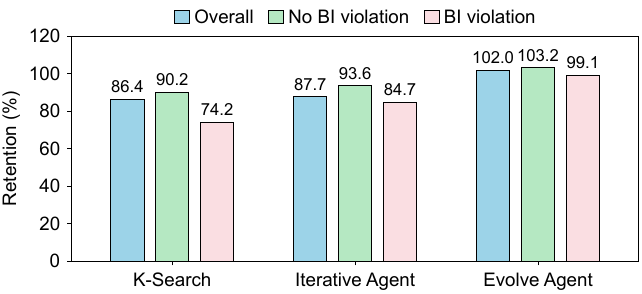}
\caption{Speedup retained by baseline BI outcome across agent frameworks. Subgroups are grouped by the baseline final kernel's BI outcome.}
\label{fig:performance_subgroups}
\end{figure}

Figure~\ref{fig:performance_subgroups} compares performance retention across baseline BI outcomes. We partition tasks according to whether their vanilla kernels have a confirmed BI violation.

Aggregate retention is lower for tasks with confirmed violations than for tasks without violations (86.0\% versus 95.7\%), consistent with BI gating excluding fast but batch-dependent optimization choices \citep{he2025nondeterminism}. The Evolve Agent retains nearly all performance in both subgroups, suggesting that broader search can recover invariant alternatives and that the BI cost depends partly on search strategy.

\begin{samepage}
\subsection{\textnormal{\tool} Component Ablation}
\label{sec:ablation}

To isolate the contribution of each \tool component from changes in the agent search trajectory, Table~\ref{tab:checker_ablation} compares static-analysis-only, dynamic-validation-only, and full \tool configurations offline on the same balanced, independently labeled candidate pool. The full \tool measurement preserves sequential early exit.

\par\end{samepage}
\begin{table}[!htbp]
\caption{\tool component ablation on a shared candidate pool ($N=200$).}
\label{tab:checker_ablation}
\centering
\small
\setlength{\tabcolsep}{4pt}
\renewcommand{\arraystretch}{1.12}
\begin{tabularx}{\linewidth}{@{}>{\hsize=1.35\hsize\linewidth=\hsize\raggedright\arraybackslash}X*{3}{>{\hsize=.60\hsize\linewidth=\hsize\centering\arraybackslash}X}>{\hsize=1.25\hsize\linewidth=\hsize\centering\arraybackslash}X>{\hsize=1.60\hsize\linewidth=\hsize\centering\arraybackslash}X@{}}
\toprule
Component & \multicolumn{3}{c}{\shortstack{BI violation\\detection (\%)}} & \shortstack{False rejection\\(\%)} & \shortstack{Median evaluation\\time (ms/candidate)} \\
\cmidrule(l{2pt}r{2pt}){2-4}
& Overall & Unique & Overlap & & \\
\midrule
Static module & 83 & 37 & 46 & 2 & 7.69 \\
Dynamic module & 55 & 9 & 46 & 0 & 2.24 \\
\tool & 92 & -- & -- & 2 & 10.21 \\
\bottomrule
\end{tabularx}
\end{table}

\tool detects 92\% of confirmed violations, compared with 83\% for the static analysis module and 55\% for the dynamic validation module. It rejects 2\% of candidates with no confirmed BI violation. Median evaluation time is 10.21~ms per candidate for full \tool, versus 7.69~ms and 2.24~ms for the static analysis module and dynamic validation module, respectively. The higher \tool detection rate supports the intended complementarity: static analysis covers modeled dependency patterns, while dynamic validation can expose realized behavior outside those rules.

\section{Related Work}
\label{sec:related_work}

In this section, we review LLM-based GPU kernel generation and complementary approaches to deterministic inference.

\paragraph{Kernel generation and optimization.} Existing work~\cite{du2026kernel, ke2026kernelzero, fan2026cuda, mai2026ave, ye2026cake, hari2026improving, guo2026specgen, xiao2026tilecas, sereda2026kforge, ran2025kernelband, wiedemann2026kernelfoundry, qu2026two, zhang2026learning, wei2025astra} combines kernel evaluation with feedback-driven generation, learned refinement, or structured search. KernelBench, introduced by \citet{ouyang2025kernelbench}, formulates kernel generation as replacing a PyTorch computation with a functionally equivalent, faster implementation, and studies repeated sampling and refinement using execution and profiling feedback. FlashInfer-Bench, proposed by \citet{xing2026flashinferbench}, extends evaluation to production LLM inference workloads through a unified representation of kernel tasks, implementations, and measurements. Its framework integrates correctness validation, performance benchmarking, and deployment. These are evaluation frameworks rather than individual agent architectures. Representative agent designs include Kevin by \citet{baronio2026kevin}, which trains multi-turn refinement through reinforcement learning with correctness and speedup rewards, and K-Search by \citet{cao2026ksearch}, which separates high-level optimization planning from code implementation. K-Search updates its search tree and optimization priorities through a co-evolving intrinsic world model informed by execution feedback. Despite their different designs, these approaches primarily emphasize correctness, performance, or broader evaluation robustness. Their validation criteria do not explicitly protect BI during optimization: numerical agreement with a reference does not establish bitwise consistency across batch configurations~\cite{sun2026kernelskill, zhu2026qimeng, li2026leap, tschand2025swizzleperf}. In this work, we focus on explicit BI protection inside the kernel optimization loop.

\paragraph{BI kernels and deterministic inference.} \citet{he2025nondeterminism} and \citet{vllm2026batchinvariance} provide specialized BI kernels or serving modes for supported implementations. Their designs control batch-dependent tiling and reduction schedules, requiring careful engineering to maintain performance tuning. In addition, they do not provide BI validation for arbitrary agent-generated candidates. \citet{yuan2025nondeterminism} systematically links changes in evaluation batch size and hardware configuration to numerical differences and divergent generated responses.

There are alternative serving designs reduce reliance on BI kernels. LLM-42, introduced by \citet{gond2026llm42}, generates candidate tokens through ordinary decoding, then replays them under a fixed-shape verification schedule and rolls back mismatches. CoRun, introduced by \citet{zhao2026corun}, instead controls execution shapes through scheduling and padding: it processes each prompt separately and pads decoding batches to a fixed size, while restricting attention reductions that would otherwise vary with batching. These methods illustrate complementary ways to obtain deterministic inference, whereas our contribution concerns protecting BI as agents optimize kernel implementations.

\section{Conclusion and Future Work}

BI should be an explicit evaluation dimension for generated GPU kernels when task semantics treat samples independently. Our analysis of released artifacts shows that conventional correctness and speed evaluation can miss batch-dependent outputs. We introduce \tool, which plugs into kernel generation frameworks, making BI actionable within existing generation and optimization workflows. On the configured test domain, \tool removes most observed violations while preserving conventional correctness and speedup. Together, these findings support evaluating BI alongside correctness and performance.

Several limitations motivate future work. The static analysis module currently provides limited cross-file analysis. Extending this analysis would improve coverage of dependencies spanning implementation boundaries. Furthermore, agent-based extraction and learning from representative examples provide paths for automatic design rule synthesis and refinement. Finally, our experiments measure batch dependence at the kernel output level, future work should assess its end-to-end effects on token generation, evaluation, serving, and reinforcement learning.

\bibliography{references, cm, llmsys}

\bibliographystyle{unsrtnat}

\clearpage
\appendix
\setcounter{table}{0}
\renewcommand{\thetable}{A\arabic{table}}
\renewcommand{\theHtable}{appendix.\arabic{table}}

\section*{Appendix Table of Contents}
\begin{itemize}
\item \hyperref[app:taxonomy]{Appendix~\ref*{app:taxonomy}: Problem evidence by kernel family and mechanism}
\item \hyperref[app:dsl_contract_example]{Appendix~\ref*{app:dsl_contract_example}: Example DSL Contract}
\item \hyperref[app:protocol]{Appendix~\ref*{app:protocol}: Task selection and evaluation details}
  \begin{itemize}
    \item \hyperref[app:coverage]{Appendix~\ref*{app:coverage}: Task coverage}
  \item \hyperref[app:protocol_details]{Appendix~\ref*{app:protocol_details}: Protocol details}
  \end{itemize}
\item \hyperref[app:late_search]{Appendix~\ref*{app:late_search}: Late-stage search gains}
\end{itemize}

\section{Problem evidence by kernel family and mechanism}
\label{app:taxonomy}
Table~\ref{tab:manual_taxonomy} complements the aggregate results in Table~\ref{tab:released_audit} by summarizing where confirmed BI violations and structural risks appear across kernel families. Together, the two tables provide evidence that batch dependence occurs in released kernels and spans multiple implementation mechanisms.

\begin{table}[htbp]
\caption{Batch invariance evidence by kernel family and mechanism. Confirmed: test confirmed BI violation with the associated mechanism identified. Risk: potential risk identified by static inspection, with no test confirmed violation. None: no evidence reported for that mechanism, not proof of absence. N/A: no semantic batch dimension in the analyzed artifacts. Mechanisms are not mutually exclusive. GEMM = general matrix--matrix multiplication.}
\label{tab:manual_taxonomy}
\vspace{\baselineskip}
\centering

\setlength{\tabcolsep}{4pt}
\renewcommand{\arraystretch}{1.12}
\begin{tabularx}{\linewidth}{@{}Xccccc@{}}
\toprule
\raisebox{0.5\baselineskip}{Kernel family} & \shortstack{Evidence\\category} & \shortstack{Work\\partitioning} & \shortstack{Reduction\\order} & \shortstack{Algorithm\\selection} & \shortstack{Expert\\routing} \\
\midrule
Attention & High & Confirmed & Confirmed & Confirmed & None \\
Linear attention & Medium & Confirmed & None & Confirmed & None \\
Mixture of experts & Low & None & Risk & Risk & Risk \\
Top-$k$ & Low & Risk & None & Risk & None \\
Recurrent & Low & None & None & Risk & None \\
Triangle multiplication & Low & None & Risk & Risk & None \\
Transformer decoding & N/A & N/A & N/A & N/A & N/A \\
GEMM & N/A & N/A & N/A & N/A & N/A \\
\bottomrule
\end{tabularx}
\par\smallskip
\end{table}

\paragraph{Evidence categories.}
We summarize the evidence as follows: High denotes a confirmed violation rate of at least 50\% among batch-applicable artifacts (batch-applicable denotes kernels whose per-sample outputs are expected to be independent of the surrounding batch context); Medium denotes at least one confirmed violation with a rate below 50\%; Low denotes no confirmed violation but at least one identified structural risk; N/A denotes no true semantic batch dimension in the analyzed artifacts. Low is an evidence category, not a measured low violation probability.

\paragraph{Mechanism definitions.}
\emph{Work partitioning} includes split-K/split-KV, sequence chunking, split--merge, recurrence segmentation, and blocks per row. \emph{Reduction order} includes partial-sum grouping, reduction-tree structure, \texttt{BLOCK\_K} or tiling changes that alter accumulation order, and partial-result merge order. \emph{Algorithm selection} includes autotuning, shape- or batch-dependent dispatch, backend kernel selection, and alternate implementation paths. \emph{Expert routing} includes expert--token grouping, variable expert sizes, and grouped-GEMM scheduling. These patterns threaten BI when a permitted change in batch context alters the computation for a preserved target sample; a pattern alone is not evidence of a violation. Here, K is the matrix reduction dimension, KV denotes keys and values, and \texttt{BLOCK\_K} is the reduction tile size. Related BI inference analyses discuss the numerical consequences of changing reduction and partition choices \citep{he2025nondeterminism}.

\section{{Example DSL Contract}}
\label{app:dsl_contract_example}

{The following listing reproduces the implemented \texttt{batch\_identity.bid} contract. The registry loads this file directly through \texttt{parse\_contract\_file()}; therefore, the listing uses executable contract syntax rather than pseudocode.}

\begin{tcolorbox}[colback=black!3,colframe=black!50!black,coltext=black]
\ttfamily
kernel BatchIdentity \{\\
\quad invariant\_axis batch symbols [B, batch, batch\_size, bsz,\\
\qquad num\_seqs, len\_indptr, num\_requests, num\_tokens, total\_q,\\
\qquad total\_kv, num\_kv\_indices, total\_pages];\\
\quad parameter NUM\_SPLITS \{\\
\qquad semantic split\_structure;\\
\qquad policy fixed\_across\_batch;\\
\qquad aliases [num\_splits, max\_splits, num\_partitions,\\
\qquad\quad split\_factor, split\_k, splits, ns, k\_split, S, \_S,\\
\qquad\quad SPLIT, SPLITS, K\_SPLIT, SPLIT\_K, NUM\_PARTITIONS];\\
\quad \}\\
\quad parameter RNG\_SEED \{\\
\qquad semantic rng\_stream;\\
\qquad policy fixed\_across\_batch;\\
\qquad aliases [seed, rng\_seed, seed\_counter, rng\_counter];\\
\quad \}\\
\}
\end{tcolorbox}

{The \texttt{invariant\_axis batch symbols} declaration defines the batch-source vocabulary. For \texttt{NUM\_SPLITS}, the \texttt{fixed\_across\_batch} policy marks the canonical parameter and every listed alias as protected sinks. The analyzer consequently rejects direct or transitive data dependence, as well as batch-derived control dependence through conditionals or loops, from any declared batch symbol to \texttt{NUM\_SPLITS} or an alias. Dependencies that cannot be traced to a batch symbol declared in the contract are outside the current static-analysis scope.}

{Here, the three protected parameters are semantic abstractions rather than exhaustive code patterns: \texttt{NUM\_SPLITS} covers batch-dependent partitioning and reduction structure across aliases and kernel families, while opaque algorithm selection, expert-routing behavior, and other undeclared runtime effects are covered by the complementary dynamic validator.}\par\section{Task selection and evaluation details}
\label{app:protocol}
\subsection{Task coverage}
\label{app:coverage}
We construct the suite to provide compact coverage of BI applicable kernel semantics rather than exhaustively enumerate FlashInfer configurations. Tasks are required to have a meaningful semantic batch dimension and are selected to cover distinct operator families and execution regimes. We prioritize higher risk attention and linear-attention kernels while retaining MoE, normalization, and sampling tasks to broaden operator coverage. Within attention, the suite spans GQA, MLA, and DSA as well as representative paged/ragged and prefill/decode regimes; linear attention covers KDA and Gated DeltaNet. We prioritize these semantic and operator axes over additional configurations that would largely duplicate the same coverage. Overall, we evaluate diverse kernel definitions covering 1,180 benchmark workloads/shapes.

\subsection{Protocol details}
\label{app:protocol_details}
\paragraph{Evaluation protocol.} We use each framework's original correctness, performance benchmarking, and failure-handling procedures \citep{cao2026ksearch,flashinfer2026agents}. Controlled generation uses Gemini 3.7 Flash for all three frameworks. For dynamic validation, we construct sample~0 at batch size one and embed the same input unchanged as sample~0 in each larger batch, comparing only its output bitwise. Each context uses one preserved target sample; testing stops at the first mismatch. The \tool workflow and acceptance gates were designed independently of evaluation data, which are used only for measurement. Dynamic validation intentionally samples a lightweight finite subset of the formal BI context space; static analysis complements it by tracing all declared batch-context symbols, not only batch size, to protected sinks. Thus, the combined modules are intended to enforce BI without exhaustive context enumeration.

\paragraph{Shared candidate pool.} We perform an \tool component ablation on one fixed pool of 200 generated candidates: 100 independently confirmed BI violations and 100 candidates with no confirmed BI violation. Candidates are drawn across the three agent frameworks, vanilla and BI-aware generation, and the major task families. Selection is independent of the observed decisions under the static-analysis-only, dynamic-validation-only, and full \tool configurations. We prioritize distinct violation mechanisms and limit near-duplicate candidates from the same task and run.

\paragraph{Independent BI labeling.} Labels are assigned independently of \tool output. Static inspection identifies candidate mechanisms and relevant batch boundaries; an independent local test then crosses the boundary and compares the preserved output bitwise. A candidate is confirmed only when this test observes an actual bitwise output difference for the unchanged target sample. Candidates are labeled as confirmed BI violation, or no confirmed BI violation; all unclassifiable candidates are excluded from the ablation pool.

\paragraph{\tool component evaluation and timing.} Static-analysis-only, dynamic-validation-only, and full \tool configurations are applied to the shared candidate pool. Violation detection is the fraction of confirmed violations rejected by the evaluated configuration. False rejection is the fraction of candidates without confirmed violations rejected by the evaluated configuration. Evaluation time excludes LLM and application programming interface (API) calls, candidate compilation and setup, reference correctness testing, and autotuning or performance benchmarking. Dynamic validation time includes all GPU kernel launches issued by the dynamic validation module and device synchronization. Each configured batch context is warmed up 3 times and then measured over 10 synchronized repetitions. We report the median across candidates; the full \tool measurement preserves its sequential execution and early exit behavior.

\section{Late-stage search gains}
\label{app:late_search}
Table~\ref{tab:late_search} reports the change in best-so-far speedup over the final five optimization turns, grouping tasks by the BI outcome of the baseline final kernel. For each subgroup and condition, gain is the relative change in geometric mean best-so-far speedup over the final five optimization turns: $100(S_{\mathrm{end}}/S_{\mathrm{start}}-1)$, where $S$ is the geometric mean best-so-far speedup across tasks. Two tasks whose final kernels failed correctness are excluded, leaving 37 tasks. Thus, Table~\ref{tab:late_search} characterizes late-stage search progress, whereas main Figure~\ref{fig:performance_subgroups} compares final speedups.

\begin{table}[htbp]
\caption{Late-stage search gains by baseline BI outcome.}
\label{tab:late_search}
\vspace{\baselineskip}
\centering

\begin{minipage}{0.85\linewidth}\centering
\setlength{\tabcolsep}{6pt}
\renewcommand{\arraystretch}{1.12}
\begin{tabularx}{\linewidth}{@{}Xccc@{}}
\toprule
\multirow{2}{*}{\shortstack[l]{BI outcome of the\\baseline final kernel}} & \multirow{2}{*}{\shortstack{Tasks,\\$n$}} & \multicolumn{2}{c}{Late-stage speedup gain (\%)} \\
\cmidrule(l){3-4}
 & & Baseline & BI-aware \\
\midrule
Confirmed violation & 9 & 6.11 & 26.30 \\
No confirmed violation & 28 & 13.87 & 13.51 \\
\bottomrule
\end{tabularx}
\par\smallskip
\end{minipage}
\end{table}

For tasks whose baseline final kernels have a confirmed BI violation, BI-aware generation achieves a larger late-stage gain (26.30\% versus 6.11\%). This pattern indicates that useful optimization continues later under BI constraints for this subgroup, but it does not establish performance beyond the fixed search budget.


\end{document}